\PassOptionsToPackage{unicode}{hyperref}
\PassOptionsToPackage{hyphens}{url}
\documentclass[
]{article}
\usepackage{amsmath,amssymb}
\usepackage{lmodern}
\usepackage{iftex}
\ifPDFTeX
  \usepackage[T1]{fontenc}
  \usepackage[utf8]{inputenc}
  \usepackage{textcomp} % provide euro and other symbols
\else % if luatex or xetex
  \usepackage{unicode-math}
  \defaultfontfeatures{Scale=MatchLowercase}
  \defaultfontfeatures[\rmfamily]{Ligatures=TeX,Scale=1}
\fi
\IfFileExists{upquote.sty}{\usepackage{upquote}}{}
\IfFileExists{microtype.sty}{% use microtype if available
  \usepackage[]{microtype}
  \UseMicrotypeSet[protrusion]{basicmath} % disable protrusion for tt fonts
}{}
\makeatletter
\@ifundefined{KOMAClassName}{% if non-KOMA class
  \IfFileExists{parskip.sty}{%
    \usepackage{parskip}
  }{% else
    \setlength{\parindent}{0pt}
    \setlength{\parskip}{6pt plus 2pt minus 1pt}}
}{% if KOMA class
  \KOMAoptions{parskip=half}}
\makeatother
\usepackage{xcolor}
\usepackage{longtable,booktabs,array}
\usepackage{multirow}
\usepackage{calc} % for calculating minipage widths
\usepackage{etoolbox}
\makeatletter
\patchcmd\longtable{\par}{\if@noskipsec\mbox{}\fi\par}{}{}
\makeatother
\IfFileExists{footnotehyper.sty}{\usepackage{footnotehyper}}{\usepackage{footnote}}
\makesavenoteenv{longtable}
\usepackage{graphicx}
\makeatletter
\def\maxwidth{\ifdim\Gin@nat@width>\linewidth\linewidth\else\Gin@nat@width\fi}
\def\maxheight{\ifdim\Gin@nat@height>\textheight\textheight\else\Gin@nat@height\fi}
\makeatother
\setkeys{Gin}{width=\maxwidth,height=\maxheight,keepaspectratio}
\makeatletter
\def\fps@ {htbp}
\makeatother
\usepackage[normalem]{ulem}
\ifLuaTeX
  \usepackage{selnolig}  % disable illegal ligatures
\fi
\IfFileExists{bookmark.sty}{\usepackage{bookmark}}{\usepackage{hyperref}}
\IfFileExists{xurl.sty}{\usepackage{xurl}}{} % add URL line breaks if available
\hypersetup{
  hidelinks,
  pdfcreator={LaTeX via pandoc}}

\title{Assessing MSDs before Introduction of a Cobot: Psychosocial Aspects and Employee's Subjective Experience}

\author{Emma CIPPELLETTI\textsuperscript{a\footnote{Contact author: \url{mailto:emma.cippelletti@univ-grenoble-alpes.fr}, \url{mailto:aurelie.landry@univ-grenoble-alpes.fr}}}, Soufian AZOUAGHE\textsuperscript{a}, Damien PELLIER\textsuperscript{c}, Aurélie Landry\textsuperscript{a, c,*}\\
\textsuperscript{a} Univ. Grenoble Alpes \-- LIP PC2S, BP 47 - 38040 Grenoble Cedex 9 - France\\
\textsuperscript{b} Univ. Grenoble Alpes \-- LIG, BP 47 - 38040 Grenoble Cedex 9 - France\\
}
\date{}

\begin{document}

\maketitle

\hypertarget{abstract}{%
\subsection{\texorpdfstring{Abstract }{Abstract }}\label{abstract}}

Musculoskeletal disorders (MSDs) are one of the main causes of work
disability (EU-OSHA, 2019; WHO, 2019). Several solutions, including the
cobotic system (EUROGIP, 2017), have been put forward to improve
unhealthy working conditions and prevent MSDs. We sought to identify the
MSD risk factors of workers on a screen-printed glass production line
prior to introduction of a cobot. We used a mixed data collection
technique: video observations and assessment of MSD risk factors by
expert ergonomists, and then self-confrontation interviews with six
production-line operators and subjective perception of risk factors. The
two types of assessment (by experts and by operators) showed that the
most demanding risk factors were physical (e.g., work posture) and
psychosocial (e.g., mental workload). Certain risk factors were viewed
differently by the experts and the operators. One question remains: How
can a cobot make work more meaningful for operators?

\sloppy
\hypertarget{keywords}{%
\subsection{Keywords}\label{keywords}}

Industry 4.0, Cobot, MSDs, Subjective experience, Meaning of work.

\hypertarget{ruxe9sumuxe9}{%
\subsection{Résumé}\label{ruxe9sumuxe9}}

Les Troubles Musculo-squelettiques (TMS) représentent la plus grande
cause d'invalidité au travail (EU-OSHA, 2019; WHO, 2019). Plusieurs
solutions sont proposées pour réduire la pénibilité du travail, ses
difficultés et prévenir les TMS, parmi elles, figure la cobotique,
c'est-à-dire l'implantation en situation de travail de robots
coopératifs (EUROGIP, 2017). Dans ce contexte, la présente étude porte
sur l'évaluation initiale des facteurs de risques de TMS dans une ligne
de production de sérigraphie sur verre avant l'implantation d'un robot
coopératif (cobot). Notre étude s'est basée sur une méthode mixte de
recueil de données~: observations filmées et analyse des facteurs de
risque de TMS par des experts en ergonomie, puis entretiens
d'auto-confrontation avec les 6 opérateurs de la ligne de production et
perception des facteurs de risques par les opérateurs. Les résultats
montrent que les facteurs de risques jugés comme les plus pénibles
correspondent à la charge physique (e.g. les postures) et à la charge
psychosociale (charge mentale) pour les 2 types d'évaluateurs (expert et
opérateur). Mais les résultats montrent que les opérateurs entre eux ont
des perceptions différentes de certains facteurs de risques. Ces
résultats posent la question de comment le cobot pourra soulager les
opérateurs des facteurs de risques au regard du sens du travail des
opérateurs.

\hypertarget{acknowledgments}{%
\subsection{Acknowledgments~}\label{acknowledgments}}

The authors would like to thank the company for participating in this
study, as well as the employees who agreed to be observed and
interviewed.

\hypertarget{funding-details}{%
\subsection{Funding details~}\label{funding-details}}

This work was supported by the French National Research Agency under
Grant AAPG ANR 2018 and within the framework of the "Investissements
d'avenir'' Grant ANR-15-IDEX-02

\hypertarget{introduction}{%
\section{1. Introduction}\label{introduction}}

Musculoskeletal disorders (MSDs) are a major health problem at work and
impose significant financial, human and societal costs on every country
(Bellemare et al., 2019). They are the second most common cause of
acquired disabilities in the world (Gallagher \& Schall, 2017) and in
Europe (EU-OSHA, 2019). In France, according to the health insurance
report (AMELI, 2020), MSDs accounted for slightly more than 87\% of all
work-related illnesses in 2017. They also incur very high economic costs
(Department of Employment, Social Affairs and Inclusion (MTEFPDS, 2015))
and are associated with absenteeism and long-term illness (AMELI, 2020;
Fjell et al., 2007; Trevelyan \& Haslam, 2001; World Health Organization
{[}WHO{]}, 2003), with inability to work (Kuorinka \& Forcier, 1995;
Leclerc et al., 2015) and with low rates of job retention (Sérazin et
al., 2013).

MSDs are caused or made worse by the environment and conditions of the
workplace (Forde et al., 2002; Roquelaure, 2018; WHO, 2003; Xu et al.,
2013). Most often they are characterized by discomfort and pain in the
joints, muscles and soft tissues (Levanon et al., 2012). They may also
involve tendinitis, nerve compression syndrome and lower back pain
(Whysall et al., 2006). MSDs take multiple forms, and their origins are
multifactorial. Epidemiological and ergonomic studies have shown three
main categories of causal factors: (1) physical or biomechanical factors
(e.g., repetitive movements, extreme or improper postures, excessive
physical effort, body vibrations); (2) organizational and psychosocial
factors (e.g., organization of tasks, pace of work, job
autonomy,\footnote{Job autonomy means being able to take initiative and
  thus take charge of a complex work situation, which is not totally
  subject to rules, and being able to self-organize (look for the best
  way of doing things to achieve a set objective) (Everaere, 2019).
  There are three types: procedural autonomy; planning autonomy; and
  criterion autonomy (Azouaghe, 2019; Breaugh, 1999).} work/rest ratio,
social support from co-workers and superiors); and (3) sociodemographic
and individual factors (e.g., age, sex, education level, medical
history, sports or recreational activities) (Bao et al., 2015; Bernard,
1997; Buckle \& Devereux, 2002; Caroly et al., 2010; EU-OSHA, 2019;
Lafranchi \& Duveau, 2008; National Research Council, 2001; Nunes, 2009;
Nunes \& McCauley-Bush, 2012; Roquelaure et al., 2011).

Because MSDs have become so widespread in the workplace, interest has
increased in assessing their risk factors in order to prevent them and
their consequences (Bellemare et al., 2019; France\textquotesingle s
Institute for Research and Security {[}INRS{]}, 2014, 2015).
Unfortunately, there are several obstacles due to the complexity and
multicausality of MSDs (Whysall et al., 2004; Landry, 2012). In
addition, treatment can be difficult. MSDs do not all have the same
etiology, they are perceived differently by different stakeholders and
there may be a lack of sufficient time, equipment and skills (Caroly et
al., 2008).

The Fourth Industrial Revolution (Industry 4.0) is providing us with
technological and ergonomic solutions to MSDs, particularly cobotics or
collaborative robotics (Bobillier Chaumon, 2021). Cobotics promises to
improve working conditions and reduce difficult and dangerous tasks, in
particular those that can lead to MSDs (CEA, 2015; Claverie et al.,
2013).

\hypertarget{cobotics-preserving-workers-health}{%
\section{2. Cobotics, Preserving Workers'
Health?}\label{cobotics-preserving-workers-health}}

Cobotics is offering a new way of organizing work (Kagermann et al.,
2013) and production processes (Lu, 2017) with a view to reaching higher
levels of operational efficiency, productivity and automation (Thames \&
Schaefer, 2016).

The promise of cobotics is multifold: economic, by increasing
flexibility and competitiveness; organizational, by providing new forms
of human-machine collaboration and by eliminating certain spatial
constraints; and ergonomic, by reducing the drudgery of working
conditions and by limiting the risks of MSDs (EUROGIP, 2017;
Moulières-Seban et al., 2017).

A cobot is a ``robotic device that manipulates objects in collaboration
with a human operator'' (Colgate et al., 1996, p. 433). For Jansen et
al. (2018), it is a robot designed to collaborate and interact
physically with humans in a shared workspace. There are four types of
human-robot collaboration: independent; simultaneous; sequential; and
supportive (Colim et al., 2021). ``The supportive scenario includes the
cases where the robot and operator work together in the same process and
workpiece simultaneously'' (Colim et al., 2021).

By introducing collaborative robots into industry, it is hoped that
workers will be less exposed to the risk factors of MSDs (Claverie et
al., 2013; Caroly et al., 2019; Villani et al., 2018). Cobots ``support
the employee in their actions and adjust their interventions to those of
the professional. It is no longer simply a substitute robot or a
mechanical aid for certain tasks'' (Bobillier Chaumon et al., 2019, p.
17). Humans and cobots thus work together to perform tasks defined by
computer programmers, organizational psychologists, ergonomists and
others involved in task definition, like managers and human resource
departments. For Norman (cited by Bobillier Chaumon et al. 2022, p. 2),
cobots ``do not only transform the capacities of an individual, they
also change the nature of the task that the person accomplishes.'' As
with any new workplace technology, cobotics can achieve the anticipated
results only if designers, computer programmers and ergonomists take
into account the employees' perception of their new ``co-workers.''
Participatory approaches are the preferred means (Galey et al., 2022) to
design human-cobot collaboration, the aim being to minimize exposure to
occupational hazards while preserving key activities for the operator
(Chahir et al., 2022; Colim et al., 2021). The main challenge is to
design safe interaction with intuitive interfaces and accurate task
allocation (Villani et al., 2018).

\hypertarget{subjective-experience-in-the-operators-assessment-of-msd-risk-factors-before-cobot-introduction}{%
\section{3. Subjective Experience in the Operators' Assessment of MSD
Risk Factors before Cobot
Introduction}\label{subjective-experience-in-the-operators-assessment-of-msd-risk-factors-before-cobot-introduction}}

The introduction of cobots into industry requires, at the design stage,
a clear division of production-line tasks between the operator and the
cobot (Cardoso et al., 2021). To that end, the tasks are studied and the
MSD risk factors assessed with a view to assigning the cobot the most
arduous tasks (e.g., weight bearing, rapid and repetitive wide-ranging
movements) and the human the most worthwhile and meaningful ones
(Cardoso et al., 2021; Isaksen, 2000; Morin, 2008; Vallery et al.,
2019). A task refers to an objective to be reached in circumscribed
conditions where variations occur (Leplat, 1989). It may be technical,
organizational or individual. In contrast, an activity "refers to the
worker, to the person or persons carrying out the work. It is what the
"operator" mobilizes and deploys in terms of subjectivity, understanding
and expertise, intelligence, and also efforts to deal, during his work,
with what has not been decided, or even with modifications of what has
been decided because it has become unsuitable in a given context"
(Guerin et al., 2021, p. 34).

According to Karsh et al. (2001), any assessment of MSDs purporting to
be holistic and systematic has to take into account the different
factors involved in work. Although more ergonomic interventions are
being made to alleviate MSDs, many of them are focused more on physical
load factors, and ``neglect'' the role of psychosocial factors (Cardoso
et al., 2021; Whysall et al., 2006). In human-robot collaboration
systems, physical ergonomics aims to improve operator posture in
real-time (Dimitropoulos, 2021; Kim et al., 2021; Lorenzini et al.,
2019; Shafti et al., 2019). Studies in this area rely on
instrument-based systems to assess the operator's physical state through
RULA (Rapid Upper Limb Assessment) and REBA (Rapid Entire Body
Assessment). Traditional observation is still used by some of the
studies reviewed by Cardoso et al. (2021). In all of these methods,
ergonomic experts observe the work activity to assess the MSD risk of
the external physical workload (Colim et al., 2020; David, 2005).

The psychosocial workload, however, also contributes to MSD risk, as
shown by Bourgeois and colleagues (2000), Coutarel and Daniellou (2011)
and Grenier-Pezé (2003). By focusing on the individual worker, and the
mechanical stress on body tissues caused by various tasks, particularly
their repetitiveness, intensity, duration and stress, we may end up
ignoring the collective and organizational dimensions of work (Buchmann
\& Landry, 2010, St Vincent et al., 2011). If a model considers only the
external physical workload, it will poorly represent the daily
on-the-job experience that develops through the social and subjective
history of workers. It is necessary to consider the difficulties they
experience to prevent harm to their health (Lasfargues et al., 2005). By
taking into account their perceptions, beliefs and knowledge, we can
better assess MSDs and prepare workers for organizational change; in
this case, introducing a cobot to the production line (Haslam, 2002;
Barrett et al., 2005).

On that basis, we sought to assess the MSD risk factors of work before
the introduction of a cobot to a screen-printed glass production line,
by using a participatory and integrative approach. In particular, we
sought to answer two questions:

\begin{itemize}
\item
  What are the existing MSD risk factors? And how are they assessed by
  the experts and by the operators?
\item
  Which tasks would the operators delegate to the cobot for future
  collaboration? Which tasks would they prefer to keep to themselves
  (because of the lower risk)?
\end{itemize}

\hypertarget{methods}{%
\section{4. Methods}\label{methods}}

\hypertarget{overall-procedure}{%
\subsection{\texorpdfstring{ 4.1 Overall
Procedure}{ 4.1 Overall Procedure}}\label{overall-procedure}}

We used a multi-stage approach (Figure 1). First, we took videos of each
workstation on the production line. Then we explained to the operator
why we had chosen that work situation (see also 4.2). The videos were
assessed by four experts, who used an APACT grid. Then, we used the
videos to conduct self-confrontation interviews with the operators.
Finally, using the results, we created scenarios to simulate the arrival
of a cobot on the production line. We will not describe the last stage
in this article.

\begin{figure}
\centering\includegraphics[width=6.06172in,height=3.32052in]{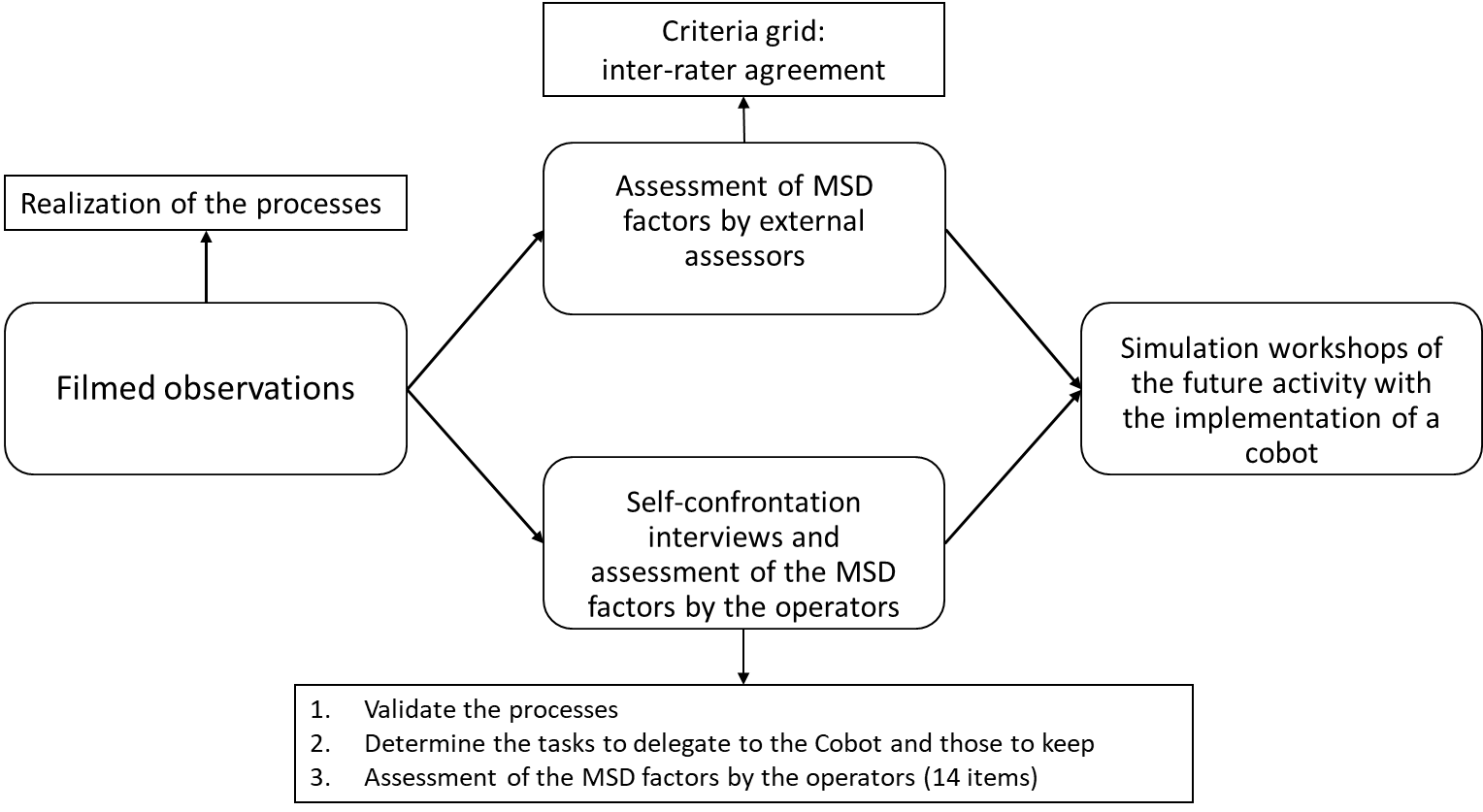}
\caption{Overall Procedure}
\end{figure}

\hypertarget{video-assessment-of-production-line-workstations}{%
\subsection{\texorpdfstring{4.2 Video Assessment of Production Line
Workstations
}{4.2 Video Assessment of Production Line Workstations }}\label{video-assessment-of-production-line-workstations}}

We made several personal observations to understand the many different
production lines. In consultation with the company CEO, we took videos
of a specific production line that corresponded to the one where a cobot
could be introduced. We filmed the workstations both at the entrance to
the oven (work processes: placing the products on the conveyor belt,
screen printing and visual inspection) and at the exit (work processes:
visual inspection, placing the products on the conveyor belt, packaging,
palletizing and preforming the cardboard boxes). There was variability
in the screen-printed products (bottles or jugs). For this article, we
will present the results only for the jugs and only for the four
workstations beyond the oven exit (Figure 2).

\begin{figure}
\centering
\includegraphics[width=3.44965in,height=3.92949in]{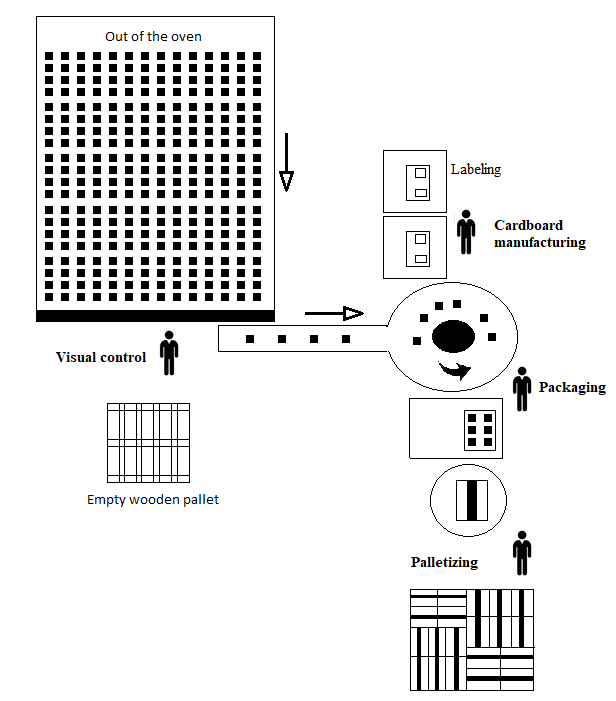}
\caption{Layout of the Four Workstations beyond the Oven Exit}
\end{figure}

The work involved taking products from the oven to a conveyor belt. The
operators (\#1 and \#2) at the ``visual inspection'' workstation would
pick up the jugs, check them and place them on a conveyor belt that led
to the packaging workstation (operator \#3). At that workstation, the
operator would pack the jugs in a cardboard box and then send the box to
the next workstation to be palletized (operators \#4 and \#5). Here, the
boxes would be taped up and put onto the pallet. The pallet-loading
station was also used to preform the cardboard boxes. The work was
repetitive, loads had to be carried, the stations were dependent on each
other and there were few areas for buffer stock. Moreover, the work
required high cognitive effort to check for the presence of product
defects, sometimes a millimetre in size. The operators signed a consent
form to show that they had agreed to participate in the study and
another consent form for image production rights. We took videos of the
three successive workstations beyond the oven door for a total of 24 min
17s.

One goal of the observations was to identify very precisely the various
actions by the operators at each workstation (type of action, movements,
sequence...) in order to provide the computer specialists with
sufficient information to start work on the movement programming of the
future cobot.

\hypertarget{assessment-by-experts}{%
\subsection{4.3 Assessment by Experts}\label{assessment-by-experts}}

To assess the MSD factors at each of the workstations under observation,
we used the APACT grid (1991), which assesses working conditions and
work organization on the basis of 22 criteria (Bernard et al., 2017).
This grid is based on traditional tools, such as RULA and REBA, to which
have been added criteria on cognitive workload and organizational
factors. It can be used to measure four risk factors: biomechanical;
psychosocial; organizational; and environmental (Jafflin \& Nadeau,
2020). We retained 14 criteria that were the most relevant to the aim of
the study and the most conducive to video assessment (Table 1). Four
experts viewed the videos to assess the workstations, using the
criteria. Two of them assessed the real work processes, and the other
two assessed the videos (during the COVID-19 lockdown). Traditionally,
MSD risk has been assessed by a single expert (St Vincent et al., 2001).
We used several experts to increase reliability (LeBreton \& Senter,
2008; Tinsley \& Weiss, 1975). Each criterion received a score between 0
and 10.5 from each expert.\footnote{The APACT grid has the following
  score ranges: 0 to 2.5: Bad; 2.5 to 4.5 Insufficient; 4.5 to 6.5:
  Medium; 6.5 to 8.5: Good; 8.5 to 10.5: Very good.} The assessments
were then pooled, and agreement on the criterion score was reached.

\hypertarget{self-confrontation-interviews-and-assessment-of-msd-risk-factors-by-the-operators}{%
\subsection{4.4 Self-Confrontation Interviews and Assessment of MSD
Risk Factors by the
Operators}\label{self-confrontation-interviews-and-assessment-of-msd-risk-factors-by-the-operators}}

\hypertarget{participants}{%
\paragraph{Participants}\label{participants}}

The sample consisted of 6 operators (3 from the morning shift and 3 from
the afternoon one)\footnote{One of the operators filmed during the first
  stage was not at work during the interviews. This person was excluded
  from calculations of the average age and seniority of the company
  operators.} who worked at the reception workstation of the production
line, where screen-printed glass jugs were coming out of the oven. There
were 2 men and 4 women, whose average age was \emph{M} = 45.8 years and
whose average seniority was 1 year and 9 months (\emph{M} = 1.75). Two
of the six operators worked for an adapted company (a business designed
for disabled workers).\footnote{``Adapted companies contribute to the
  development of territories and promote an inclusive economic
  environment favourable to disabled men and women'' (Légifrance, 2020).}
The researchers were not told about the workers' disabilities, which did
not impede communication or understanding. To ensure we were clearly
understood, we asked the operators' monitor (an employee support
position) to explain the interview procedure to them.

\hypertarget{procedure}{%
\paragraph{Procedure}\label{procedure}}

The self-confrontation interviews lasted 2 hours and 43 minutes for an
average of 32 minutes each. In a simple self-confrontation interview, an
employee would be confronted with the video of his/her activity and
asked to comment on it (Boubée, 2010; Clot et al., 2000). The operators
were shown short sequences of their activities,\footnote{The videos
  shown to the operators were the same as those used by the experts to
  assess MSD risk factors.} asked questions and prompted throughout the
viewing. The interviews were filmed. We carried out self-confrontation
interviews primarily to validate the work processes at each workstation
(e.g., a fine and detailed description of all the stages and all the
stage sequences) and secondarily to understand some aspects of the
activity (e.g., a glance toward a co-worker, a momentary stop, the
buffer stock). By analyzing the videos and the work processes, we hoped
to achieve a "participant/researcher co-analysis" (Boubée, 2010, p. 3).
To assess subjective difficulty, we asked the operators to rate 14
criteria based on physical, organizational and psychosocial factors on a
10-point Likert scale derived from the APACT grid (Table 1). To
understand what gave meaning to them in their tasks, and to investigate
a possible division of tasks between human and cobot, we asked four
questions: ``If you had the possibility tomorrow of dropping one of
these tasks, which one would it be? Conversely, which one would you
prefer to keep? Imagine that you were offered a chance to have a cobot
at the xxx workstation, which task would you wish it to do? Which task
would you wish to keep for yourself?''\newpage

\begin{table}
\begin{center}
\includegraphics{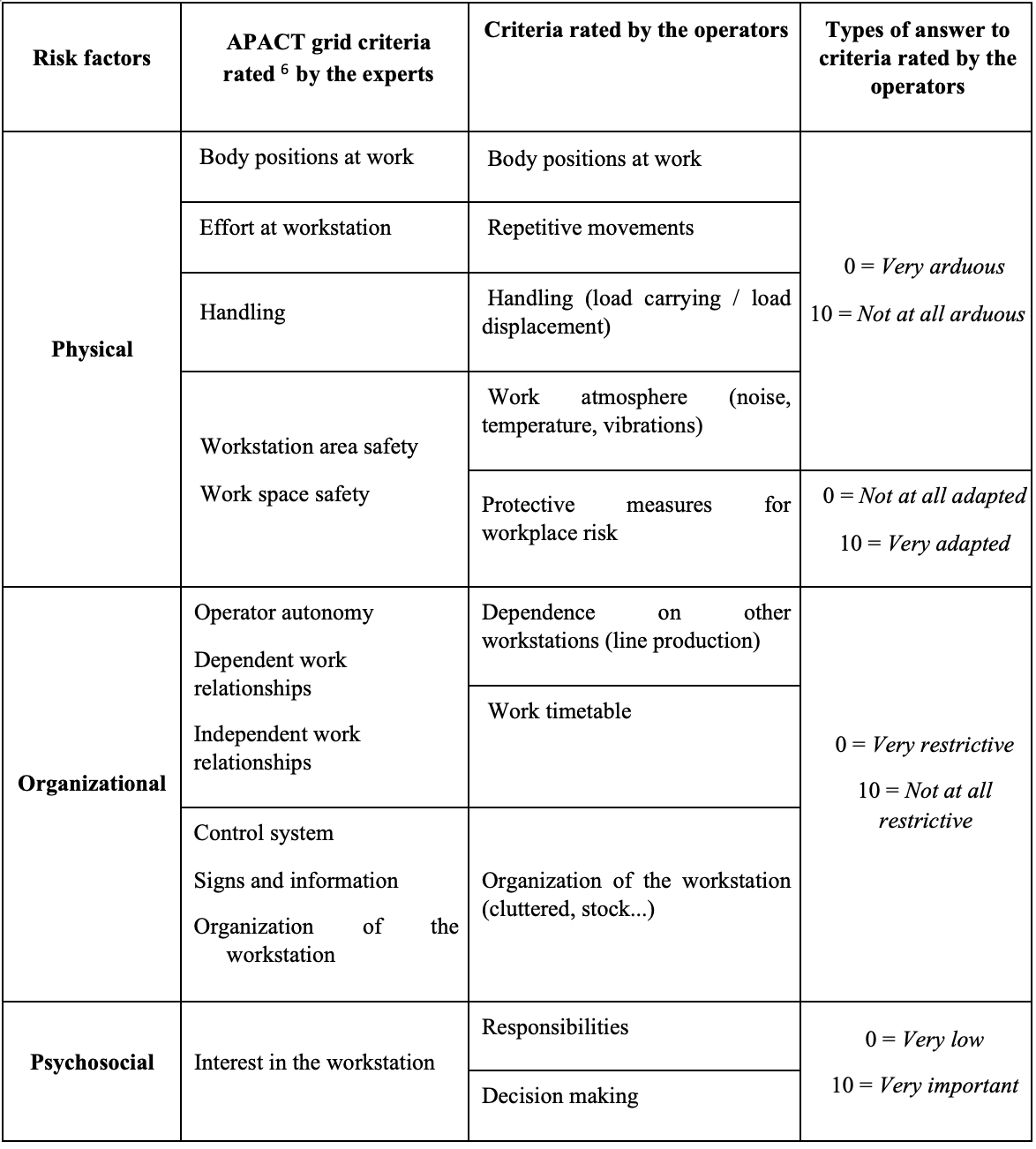}
\caption{Difficulty Level Criteria - Experts and Operators}
\end{center}
\end{table}

\section{5. Results}
\label{results}

\hypertarget{assessment-of-msd-risk-factors-by-the-experts}{%
\subsection{5.1 Assessment of MSD Risk Factors by the
Experts}\label{assessment-of-msd-risk-factors-by-the-experts}}

First, we identified the work processes by dividing the operator's task
into action units (Theureau, 2004).

According to the experts, the work situations had MSD risks. Their
assessment shows that 13 factors out of 14 were below the ``correct''
score of 6.5, as defined by the APACT grid (Figure 3). The three ``bad''
factors included two psychosocial ones (mental workload and monotony)
and a physical one (handling). Five organizational factors were rated
``insufficient'' (i.e., workstation organization, operator autonomy,
dependent and independent relationships, control system), as was one
physical factor (i.e., workstation area safety). In addition, three
physical factors (i.e., body position at work, work space safety,
efforts at the workstation) and one psychosocial factor (i.e., interest
in the workstation) were rated ``average.'' Only the signs and
information criteria were rated ``very good.''

\begin{figure}
\centering\includegraphics[width=5.32824in,height=2.77222in]{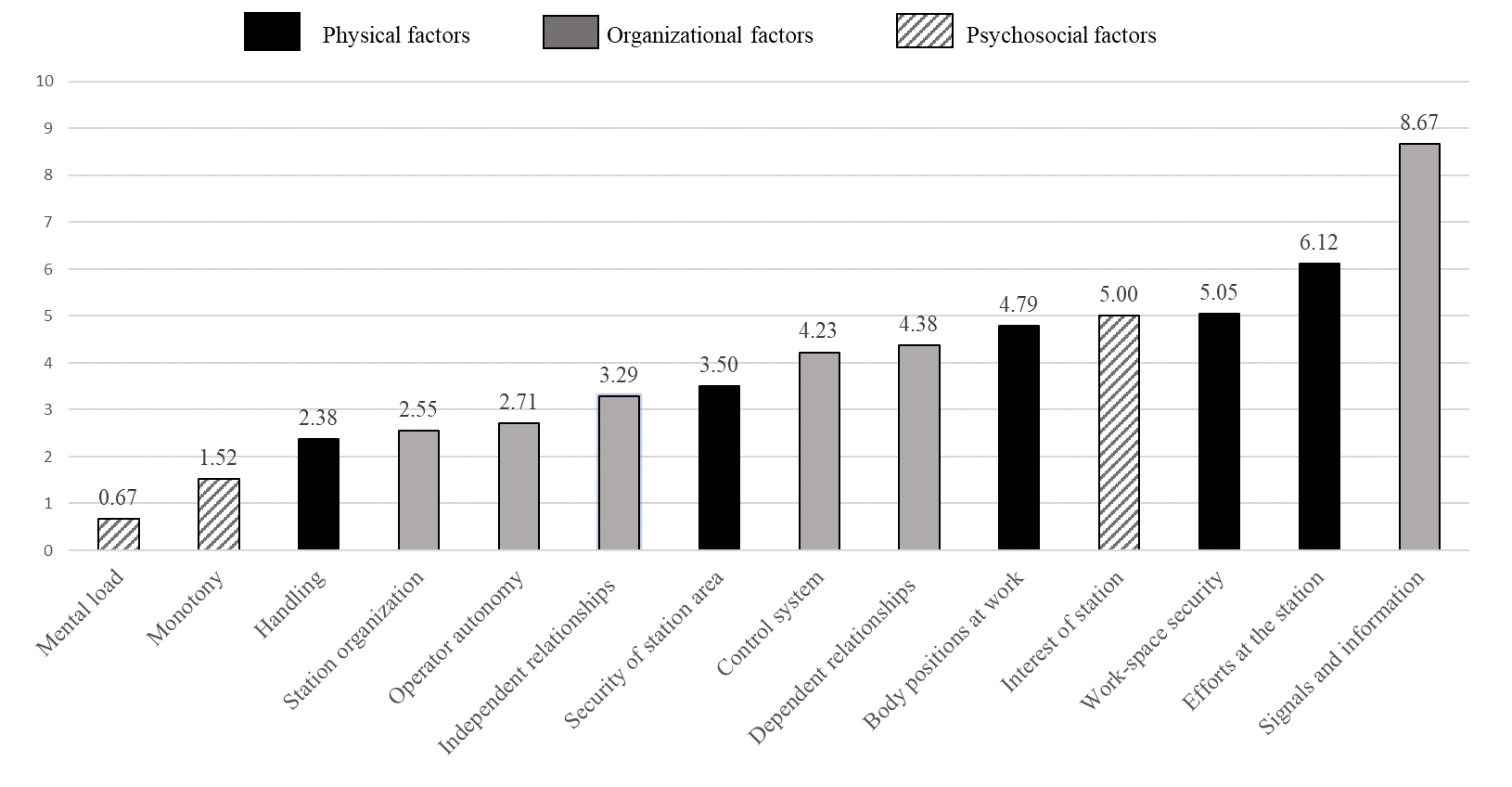}
\caption{Average Ratings of MSD Risk Factors by the Experts
(Criteria on X Axis and Ratings on Y Axis)}
\end{figure}

Although the experts agreed that the work situations were characterized
by several MSD risk factors, we asked the operators to perform their own
assessment.

\hypertarget{ratings-of-msd-risk-factors-comparing-the-operators-and-the-experts}{%
\subsection{5.2 Ratings of MSD Risk Factors: Comparing the Operators
and the
Experts}\label{ratings-of-msd-risk-factors-comparing-the-operators-and-the-experts}}

The self-confrontation interviews began with a discussion with the
operators about the different work processes. Then we analyzed their
perceptions of the 14 risk factors, asking them to choose the three that
they considered to be the most difficult in their job. We gave 3 points
to the first one they mentioned, 2 to the second and 1 to the third. In
general, the operators rated only 4 criteria ``below average'' on the
Likert scale (Figure 4). The four risk factors judged to be the hardest
were a psychosocial one (i.e., pace of work) and three physical ones
(i.e., body position at work, work atmosphere and repetitive movements).

\begin{figure}
\centering\includegraphics[width=6.08958in,height=2.8999in]{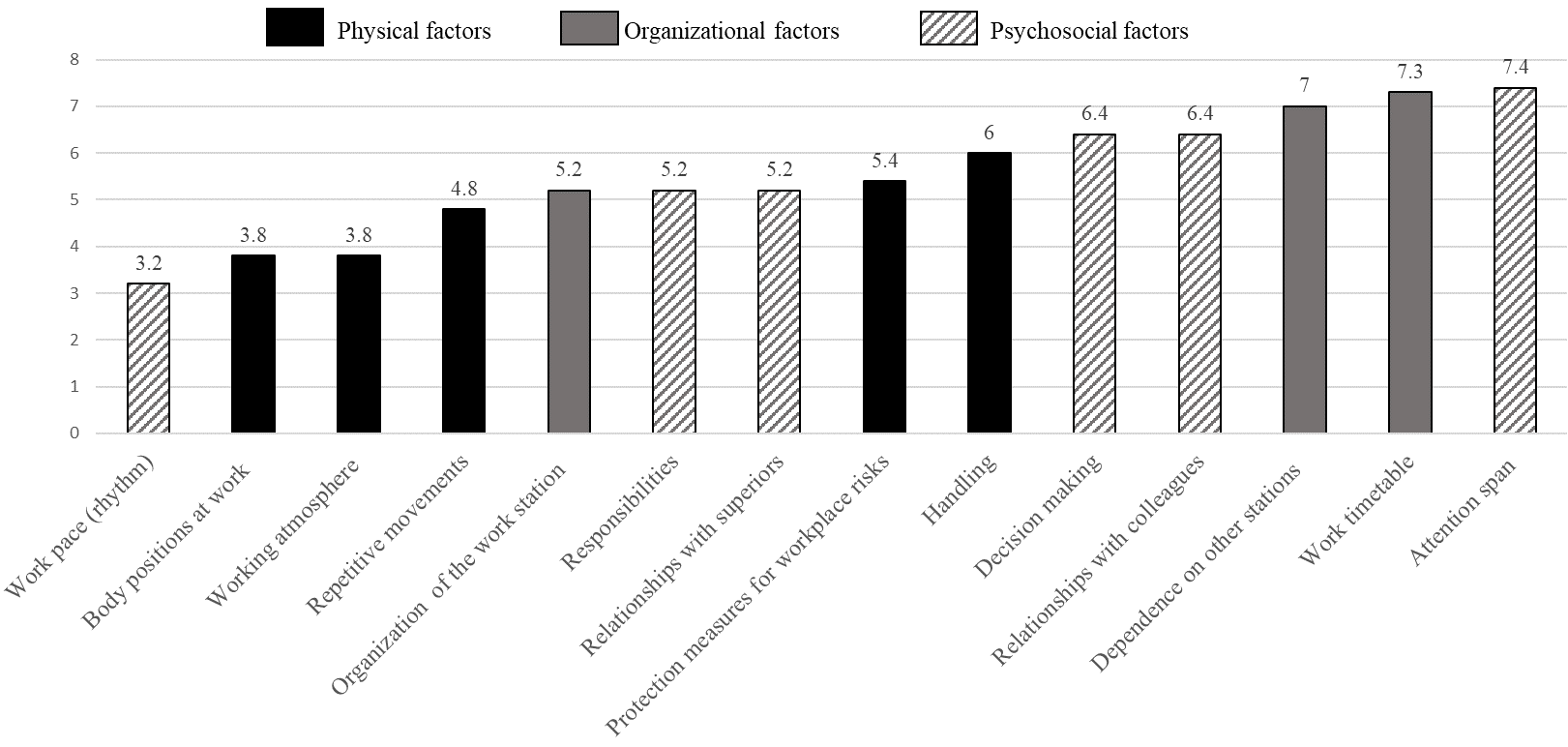}
\caption{Average Ratings of MSD Risk Factors by the Operators
  (Criteria on X Axis and Ratings on Y Axis)}
\end{figure}

Most of the highest-rated risk factors were psychosocial and physical
(Table 1). The operators considered the pace of work to be the hardest
one, and the experts similarly considered mental workload and monotony
to be the two hardest ones.

The operators also reported that the body positions for the tasks were
painful. That risk factor partly overlapped with what the experts
considered to be the third hardest risk factor: ``handling'' (which
takes the most difficult body positions into account). Nonetheless, the
experts considered ``body position at work'' to be only the tenth
hardest criterion, with a rating of 4.79. The operators rated the work
atmosphere as being the third hardest risk factor, which corresponded to
``workstation area safety'' for the experts, who rated it seventh
hardest. This risk factor had an average rating of 3.5 on the APACT
grid. It was therefore rated ``bad'' and in need of improvement.

We then analyzed the interviews in terms of all three risk factors:
physical; organizational; and psychosocial:

\uline{Physical MSD risks:} In their experience and opinion, the
operators were fairly homogeneous in assessing the body positions and
the pace of work, which they considered to be difficult:
``\emph{It\textquotesingle s true that it\textquotesingle s physical:
you have to bend down.'' ``Conditioning hurts the shoulders. You have to
go and get the cardboard at the bottom, and then it\textquotesingle s
very energetic too. You have to be energetic to do it.
It\textquotesingle s quite intense work in general. It goes quite
fast.'' ``The height of the posts is very painful. After two hours your
back hurts.''}

Several operators mentioned pallet changing as being difficult,
especially when the operators were alone.

Some results were surprising. The operators did not share the same
perception of the repetitiveness of their movements: \emph{``For me, not
at all painful. If it is to repeat the same actions, the same
movements\ldots{} That\textquotesingle s all we do anyway...''}

For some operators, this was one of the most difficult aspects of the
work, while for others it was not at all difficult.

Finally, the physical work environment (e.g., noise, heat) and the
handling were perceived overall negatively, and the means of protection
against occupational hazards were perceived overall as adequate.

\uline{Organizational MSD risks}: 7 operators positively perceived the
schedules, the organization of their workstations and the dependence of
their workstations on each other. This dependence nonetheless required a
great deal of coordination between the operators: "\emph{Depending on
the person, whether he or she goes fast or not, we are not all equal in
this respect. There are some who stockpile. There are others who do not
stockpile. There are some who send much faster than others. The aim is
to coordinate with the person who sends because afterwards we are
quickly overwhelmed.}"

In addition, most of the operators mentioned mutual aid as an integral
part of their work: "\emph{It\textquotesingle s a team job, so normally
we have to help each other. If there\textquotesingle s something wrong
or if the co-worker is having trouble, that\textquotesingle s it, we
have to help each other. {[}...{]}" "Yes, I look to see if my co-worker
has too much. I check to be able to stockpile if he
doesn\textquotesingle t make it. "I look at my co-worker. I have to
watch over the packaging because my co-worker also has to keep up. We
also have to look after our co-workers}.''

\uline{Psychosocial risks}: Most of the operators positively perceived
the responsibilities, the decision making and the relationships with
their co-workers and with the chain of command. They differed, however,
in their perceptions of visual inspection. Although both of the
operators in charge of visual inspection described it as their main
responsibility, one of them perceived it as very low in importance:
"\emph{The responsibility is very low. Maybe because we check and
that\textquotesingle s it.}" In contrast, the other operator perceived
it as important: "\emph{The responsibility is mostly {[}one of{]}
inspection. Let's say I give {[}it{]} a grade of 6. There is an
inspection, and it has to be checked. It\textquotesingle s important.}"

In the same way, two operators in the same position had totally
different perceptions of the level of attention they needed to do the
work: "\emph{Not at all difficult. I don\textquotesingle t need to be
concentrated." "For me I need to be concentrated. I
shouldn\textquotesingle t be disturbed. I should be concentrated
normally.}"

Consequently, the experts and the operators were both similar and
different in their perceptions of MSD risk factors. This point will be
further discussed in section 6. Moreover, the operators seemed to rate
only a few of the risk factors negatively. They also differed in their
perception of the drudgery of their work, thus raising the issue of its
meaningfulness to them. This factor could influence their subjective
experience of the degree of difficulty.

\hypertarget{meaning-of-work-to-delegate-or-not-to-delegate-to-the-cobot}{%
\subsection{5.3 Meaning of Work: To Delegate or Not to Delegate to
the
Cobot}\label{meaning-of-work-to-delegate-or-not-to-delegate-to-the-cobot}}

When the operators were asked which tasks they would like to delegate to
the cobot, they gave three types of answer (Figure 5). First, operators
\#2 and \#4 would have the cobot do the physical workload (handling).
They would in fact give it all of the handling, whether the product was
being moved straight out of the oven or was at the pallet-loading
station. As for operator \#3, he wished to continue handling the jugs
but have the cobot assist him, while not being clear about the kind of
assistance. He wished to keep the handling so that he could continue
looking at the jugs with a critical eye when putting them into the
cardboard boxes. It was thus the quality of the work and the visual
inspection that made him want to keep that task. The other two operators
(\#1 and \#5) wished to delegate the mental workload (visual inspection)
to the cobot. Both of them worked for an adapted company and did not
perceive the repetition of movement as a source of difficulty. The first
operator (\#5) expanded on the difficulty of maintaining his
concentration for this task over a whole day ``\emph{If it's about
putting the jugs in the box, it becomes automatic, and {[}there's{]} no
need to think. We're in automated mode.}'' The second operator (\#1)
explained that he had trouble making decisions and taking
responsibilities (these two tasks were the ones he perceived as being
the hardest in his job). He did not consider the repetitiveness of his
actions to be hard: ``\emph{for me, not hard at all. If it means
repeating the same movements, the same gestures, well, no problem at all
for me. Repeated movements, doing the same job, it doesn't bother me.}''

\begin{figure}
\begin{center}
\begin{longtable}[]{@{}
  >{\raggedright\arraybackslash}p{(\columnwidth - 4\tabcolsep) * \real{0.2185}}
  >{\raggedright\arraybackslash}p{(\columnwidth - 4\tabcolsep) * \real{0.4481}}
  >{\raggedright\arraybackslash}p{(\columnwidth - 4\tabcolsep) * \real{0.3334}}@{}}
\toprule()
\endhead
\textbf{Tasks they wanted to give to the future cobot} &
\textbf{Interview excerpts} & \textbf{Tasks they wanted to keep for
themselves} \\
Physical workload & Op \#4: Let it do the taping up and placing on the
pallets.

Op \#2: Yes, that's it: take the jugs and put them on the conveyor belt.
& Packaging workstation \\
Practical help & Op \#3: Let it hand me the boxes. It's not the end of
the world. Don't ask me, apart from that I wouldn't know. Me, I like
putting the jugs in the boxes, at the same time having a look to see if
any of them are chipped, whatever. For example, if it was the robot that
has to put them in, it wouldn't see that a lip is missing or anything
else. & \multirow{2}{*}{Handling} \\
Mental workload & Op \#1: For me, if I don't have to check what's coming
out of the oven, OK, that is one job less to do! We skip the check and
put it straight down automatically. That would suit me fine!

Op \#5: What I'd like it to do is to do the checking. That's the worst
part of the job, the most demanding. If there is a little mark, even a
tiny one, it's not good and it has to be rubbed out\ldots{} and when
we've got two seconds to do that and only two---I'm being polite---have
you just seen all those jugs and everything that has to be examined? \\
\bottomrule()
\end{longtable}

\caption{Division of Tasks between the Future Cobot and the Operators}
\end{center}
\end{figure}

Of the five operators interviewed, three said they enjoyed working at
the packaging workstation and would like to stay there. Yet it was a
workstation where the operators could not let the stock pile up if the
pace got too fast. Only the person before them (at the oven) could
stockpile on a side pallet to manage the flow of jugs. Moreover, most of
the operators said that mutual assistance was indispensable to getting
the work done properly because the different workstations were largely
dependent on each other (Figure 6). The pace of work at one station
would determine the pace of work at the next.

\begin{figure}
\begin{center}

\begin{longtable}[]{@{}
  >{\raggedright\arraybackslash}p{(\columnwidth - 0\tabcolsep) * \real{1.0000}}@{}}
\toprule()
\endhead
Operator \#4: \emph{Yes, when she's rushed off her feet or whatever, we
give each other a hand.}

Operator \#1: \emph{I look at my co-worker. I have to keep an eye on her
at the packaging station because she has to follow the pace. We have to
watch out for the co-workers. Down there, there's a co-worker who packs
the boxes, and she has to keep up with me, see?}

Operator \#2: \emph{Yes, I look to see if it's not piling up too much
for my co-worker. I check so as to slow down if he can't keep up.} \\
\bottomrule()
\end{longtable}
\caption{Excerpts from Operator Interviews}
\end{center}
\end{figure}

The results show that the operators did not necessarily wish to give a
cobot the physically hardest tasks. For some operators, the mental
workload of the visual inspection was more demanding and difficult than
the handling of the jugs.\\

\hypertarget{discussion}{%
\section{\texorpdfstring{6. Discussion
}{Discussion }}\label{discussion}}

\hypertarget{assessment-of-msd-risk-factors}{%
\subsection{6.1 Assessment of MSD Risk
Factors}\label{assessment-of-msd-risk-factors}}

For both the experts and the operators, the leading MSD risk factors
were either psychosocial or physical. As specified by Jaffar and
colleagues (2011), the physical risk factors are the most important
contributors to MSDs. The experts differed from the operators on some
points, perhaps partly because not all of them did their ratings
on-site. Nevertheless, the operators offered a subjective viewpoint that
seems important in and of itself (Lasfargues et al., 2005). Although
outside experts can definitely assess work situations and come to
conclusions, they do not necessarily identify the ones that workers
experience as being the most demanding physically or psychosocially.

The operators themselves had very different perceptions of MSD risk
factors, especially when rating repetitive movements. Hélardot (2008)
explained ``that there is no job or task that is intrinsically or
absolutely hard, but the arduousness is always relative to the
individual who has had the experience of it: it is predicated on the
state of his health, his past, his personal norms and values'' (p. 7).
Working conditions, and their accompanying difficulties, are a
subjective experience and, as such, reflected in the meaning the
operators give to their work.

\hypertarget{meaning-of-work-and-repetitiveness-of-tasks}{%
\subsection{\texorpdfstring{6.2 Meaning of Work and Repetitiveness of
Tasks
}{6.2 Meaning of Work and Repetitiveness of Tasks }}\label{meaning-of-work-and-repetitiveness-of-tasks}}

Some operators did not want to delegate the handling of jugs to the
cobot, preferring to give it either the mental workload or the visual
inspection. These results can be explained by Isaksen's research (2000),
which shows that meaning may be found in repetitive work. By including a
subjective rating of MSD risk factors, we can ask questions about the
operators\textquotesingle{} preferences, in particular what it is about
their work that enables them to maintain their health despite the
drudgery (Bakker \& Demerouti, 2017, 2018). Activity is not only about
exposure to risk factors but also about finding ways to achieve the
goals of a task while maintaining opportunities for action. These
"resources" must be studied during cobot introduction so as not to
deprive work of its meaning. They can be perceived and identified only
by the operators. To our knowledge, there is no grid that an external
observer can use to quantify and qualify such resources. This probably
explains not only the rating differences between the operators and the
experts but also the rating differences among the operators. Indeed,
resources may or may not be identifiable to employees. In a
developmental approach to design, it is thus necessary to consider how
the employees view their work and its perceived drudgery.

There are broader issues at stake here than prevention of occupational
hazards. There are also many human resource issues: skills development;
career path support; and organizational determinants of employee health.

It is important to note that the two operators who wished to keep the
handling tasks were employed by an adapted company. There has been
little research on occupational hazard assessment by workers with
special cognitive needs (Groizeleau et al., 2019; Guyon Taillens et al.,
2020). They may not attach the same meaning to work. Guyon Taillens and
colleagues (2020) explain the difficulties in assessing psychosocial
risk with workers who have special needs. They may perceive some factors
like autonomy---normally an advantage at work---as a threat and a source
of risk.

Bobillier Chaumon and colleagues (2019) argue that to gain acceptance
for emerging technologies, such as collaborative robots, one must show
their usefulness and ease of implementation for an activity. Thus, as
part of the activity, emerging technologies would not only give meaning
to the operators ``but also give (again) meaning to the activity: by
maintaining the workers' power to act, by developing their skills, by
recognizing their know-how, by increasing initiative and autonomy and by
giving a new dynamic to the job'' (p. 21). Thus, when designing the
tasks and dividing them between the cobot and the operator, one should
consider not only the different perceived difficulties of the tasks but
also the resources that the work situation offers. Because different
workers may have different views, one should prevent future problems by
distributing the tasks through a collective and participatory approach,
without omitting the pragmatic realities of programming the cobot.

\hypertarget{cobot-operator-collaboration}{%
\subsection{6.3 Cobot-Operator
Collaboration}\label{cobot-operator-collaboration}}

During the interviews, the operators said that mutual assistance was
central to their work. Because they worked on a production line, the
workstations were by definition very dependent on each other. This
system of mutual assistance could be disrupted by cobots
(Dubreuil-Nayrac et al., 2019). Bobillier Chaumon et al. (2019) report
that with the new forms of interaction inherent to these emerging
technologies ``new forms of man-robot cooperation and interfaces are to
be imagined and developed'' (p. 17). Moreover, the cobot has to
understand and adapt to the worker's intentions (Devy, 2012). The
question here is to know how a cobot can understand and anticipate the
intentions of a worker who is regulating the work flow for the sake of a
co-worker at the next workstation down the line (Ferreira Duarte et al.,
2018).

To design a cobot that can reduce the drudgery of work, it is important
to consider two points:

\begin{itemize}
\item
  First, the objective drudgery is not always the experienced drudgery.
  The operators do not always wish to delegate the worst tasks to a
  cobot.
\item
  Second, there is variability among individual operators. This point
  must be taken into account when a cobot is brought into a production
  line.
\end{itemize}

Our research shows the differences in the way a task may be perceived
within the same group of workers. Such differences will affect
constraints and resources. Because of this study, we began collaborating
with programmers to design algorithms that allow for variability not
only among individuals but also among technologies and among
organizations. This is how artificial intelligence should be adapted to
the real world of work if it is to develop. For now, it is still too
rigid and little used in industry.

\hypertarget{study-limitations}{%
\subsection{\texorpdfstring{6.4 Study Limitations
}{6.4 Study Limitations }}\label{study-limitations}}

Like any study, this one has a certain number of limitations. First, the
number of participants was limited. On the other hand, all the operators
were specialized in work on the same production line, and the
company\textquotesingle s workforce was not large. Second, some bias may
have been present in one way or another during data collection, during
the observation stage (e.g., selection effect, observer bias) and during
the self-confrontation interviews (e.g., social desirability bias). The
experts would have been more comparable with each other in their ratings
if they had all done the ratings on-site, rather than by video, as was
the case with half of them (e.g., principle of compatibility).

\hypertarget{contribution-and-conclusion}{%
\section{7. Contribution and
Conclusion}\label{contribution-and-conclusion}}

We were able to explore our initial analyses with the company CEO and
with the team of computer engineers in charge of cobot programming. We
are now left with two key questions.

First, what role ``must'' the cobot have in the work situation? Indeed,
the ``new co-worker'' must not deprive the operators of the meaning they
give to their work by taking on all the tasks that may make the
operator's job meaningful, be they quality ones or hard ones.

Second, how will the cobot be integrated into the operators' system of
mutual assistance? It will also have to anticipate and adapt itself to
the production-line operator it is there to assist. It should help not
hinder.

The introduction of a cobot can raise real Quality of Life at Work (QLW)
issues. Simultaneous human-robot collaboration calls into question the
distribution of tasks, the focus of the work activity and the perceived
drudgery. Such technological change must be assisted by HR managers,
work psychologists or ergonomists. Introduction of a collaborative robot
may also require training courses to teach the workers new skills and to
help them rethink their career paths. Technological change alone should
not be seen as the cure for all work-related ills.

We adopted a participatory approach of asking the operators to assess
the MSD risk factors, and this approach may be used by all kinds of
employees, including people with or without special needs. Different
employees may differ, however, in the way they perceive the meaning of
work, and this reality must be understood and integrated into any new
technology, like the cobot, not only by the company directors but also
by those who design the new technology (Bachellerie et al., 2022; Galey
et al., 2022).

We wish therefore to pursue our research in three ways. First, we would
like to develop a repository for data on employee variability before
cobot introduction (Fournier et al., 2023) and a method for integrating
the meaning of work into the design of human-robot collaboration.
Second, it is important to develop and validate a grid for subjective
assessment by workers of MSD risk factors. Third, we wish to develop a
cobot acceptability questionnaire for manufacturers to investigate
cobot-related perceptions, beliefs and attitudes (Cippelletti et al.,
2023) in order to gain insight not only into those factors that may
hinder the introduction of cobots but also into those that may help
integrate them into the production process.

\textbf{Disclosure statement}

No potential conflict of interest was reported by the author(s).

\hypertarget{references}{%
\section{8. References}\label{references}}

AMELI (2020) Définition et impact des TMS.
https://www.ameli.fr/entreprise/sante-travail/risques/troubles-musculosquelettiques-tms/tms-definition-impact

APACT (1991). \emph{Guide d\textquotesingle évaluation des conditions et
organisations du travail}. Paris: Association de la prévention des
Conditions de Travail.

Azouaghe, Soufian. (2019). \emph{Santé psychologique au travail dans le
milieu scolaire public : étude des déterminants organisationnels et
psychologiques chez les enseignants marocains} (Doctoral dissertation,
Université Grenoble Alpes).
\href{https://theses.hal.science/tel-02371977v1}{\uline{https://theses.hal.science/tel-02371977v1}}

Bachellerie, Camille, Gaudart, Corinne, \& Petit, Johan. (2022).
L'analyse des usages de technologies digitales dédiées à la
synchronisation : intérêts pour l'étude d'une transformation
organisationnelle dans l'ingénierie de conception automobile.
\emph{Relations industrielles / Industrial Relations}, \emph{77}(3).
https://doi.org/10.7202/1094213ar

Bakker, Arnold B., \& Demerouti, Evangelia (2017). Job
Demands--Resources Theory: Taking Stock and Looking Forward.
\emph{Journal of Occupational Health Psychology, 22}, 273-285. doi:
10.1037/ocp0000056.

Bakker, Arnold B., \& Demerouti, Evangelia (2018). Multiple levels in
job demands-resources theory: Implications for employee well-being and
performance. In: E. Diener, S. Oishi, \& L. Tay (Ed.), \emph{Handbook of
well-being}. Salt Lake City, UT: DEF Publishers. doi:
\href{http://nobascholar.com}{nobascholar.com}.

Bao, Stephen S., Kapellusch, Jay M., Merryweather, Andrew S., Thiese,
Matthew S., Garg, Arun, Hegmann, Kurt T., \& Silverstein, Barbara A.
(2015). Relationships between job organizational factors, biomechanical
and psychosocial exposures, \emph{Ergonomic}s, 59(2), 179-193,
\href{https://doi.org/10.1080/00140139.2015.1065347}{\uline{https://doi.org/10.1080/00140139.2015.1065347}}.

Barrett, Jackie H., Haslam, Roger A., Lee, Katherine G., \& Ellis, Mike
J. (2005). Assessing attitudes and beliefs using the stage of change
paradigm---case study of health and safety appraisal within a
manufacturing company. I\emph{nternational Journal of Industrial
Ergonomics}, 35, 871-887. https://doi.org/10.1016/j.ergon.2004.12.004.

Bellemare, Marie, Caroly, Sandrine, \& Prud'homme, Daniel (2019).
Travail collectif pluridisciplinaire dans la prévention des risques
professionnels complexes: ressources et contraintes du contexte au
Québec et en France. \emph{Relations industrielles/Industrial
Relations}, 74(2), 242-265.

Bernard, Bruce (Ed. 1997). Musculoskeletal Disorders and Workplace
Factors: A critical review of epidemiologic evidence for work-related
musculoskeletal disorders of the neck, upper extremity, and low back.
https://www.cdc.gov/niosh/docs/97-141/pdfs/97-141.pdf?id=10.26616/NIOSHPUB97141.

Bernard, Fabien , Sagot, Jean-Claude, \& Paquin, Raphael (2017).
Simulation de l\textquotesingle activité de maintenance pour une
meilleure intégration du facteur humain en maintenabilité. Papier
presented at: 24éme colloque des Sciences de la conception et de
l\textquotesingle innovation, CONFERE 2017, Seville, Spain.

Bobillier Chaumon, Marc-Éric, Barville, Nadia, \& Crouzat, Pascal
(2019). Les technologies émergentes au travail. \emph{Le Journal Des
Psychologues}, \emph{n°367}(5), 16--21.
\href{https://doi.org/10.3917/jdp.367.0016}{\uline{https://doi.org/10.3917/jdp.367.0016}}.

Bobillier Chaumon, Marc-Éric (2021). Technologies émergentes et
transformations digitales de l'activité\,: Enjeux pour l'activité et la
santé au travail. \emph{Psychologie du Travail et des Organisations,}
27(1), 17‑32. https://doi.org/10.1016/j.pto.2021.01.002

Bobillier Chaumon, Marc-Eric, Delgoulet, Catherine, Greenan, Nathalie,
Lemonie, Yannick, \& Warhurst, Chris (2022). Éprouver la dualité des
technologies digitales en croisant les regards disciplinaires /
Cross-disciplinary perspectives on the duality of digital technologies.
\emph{Relations industrielles / Industrial Relations}, \emph{77}(3).
https://doi.org/10.7202/1094207ar

Boubée, Nicole (2010). La méthode de l'autoconfrontation: une méthode
bien adaptée à l'investigation de l'activité de recherche d'information
?. Études de communication, 35, 47-60.

Bourgeois, Fabrice, Lemarchand, Claude, Hubault, François, Brun,
Catherine, Polin, Alexis, \& Faucheux Jean-Marie (2000). \emph{TMS et
travail, quand la santé interroge l'organisation.} Lyon: Editions de
l'ANACT.

Breaugh, James A. (1999). Further Investigation of the Work Autonomy
Scales: Two Studies. \emph{Journal of Business and Psychology, 13},
357-373. doi: 10.1023/A:1022926416628.

Buchmann, Willy, \& Landry, Aurélie (2010) Intervenir sur les TMS : Un
modèle des Troubles Musculosquelettiques comme objet intermédiaire entre
ergonomes et acteurs de l'entreprise ?. \emph{Activités}, 7(2), 84-103.

Buckle, Peter, \& Devereux, Jason (2002). The nature of work-related
neck and upper limbmusculoskeletal disorders. \emph{Applied Ergonomics,}
33, 207-217.
\href{https://doi.org/10.1016/S0003-6870(02)00014-5}{\uline{https://doi.org/10.1016/S0003-6870(02)00014-5}}.

Cardoso, André, Colim, Ana, Bicho, Estela, Braga, Ana Cristina, Menozzi,
Marino, \& Arezes, Pedro (2021). Ergonomics and Human Factors as a
Requirement to Implement Safer Collaborative Robotic Workstations: A
Literature Review. \emph{Safety}, 7(4), 71. MDPI AG. Retrieved from
http://dx.doi.org/10.3390/safety7040071

Caroly, Sandrine, Coutarel, Fabien, Escriva, Evelyne, Roquelaure, Yves,
Schweitzer Jean-Marieu, \& Daniellou, François (2008). La prévention
durable des TMS : Quels freins ? Quels leviers d'action ?
https://halshs.archives-ouvertes.fr/halshs-00373778/document.

Caroly, Sandrine, Coutarel, Fabien, Landry, Aurélie, \& Mary-Cheray,
Isabelle (2010). Comparison of production management for continuous
improvement and the OHS management system for safety and health of
workers in assembly lines sector. \emph{Applied Ergonomics} 41 (2010)
591--599.

Caroly, Sandrine, Hubaut, Rémi, Guelle, Kévin, \& Landry, Aurélie
(2019). Le travail digital, un enjeu pour les psychologues du travail.
\emph{Le Journal des psychologues}, (5), 27-32.

CEA (French Alternative Energies and Atomic Energy Commission). (2015).
\emph{Les robots s'intègrent dans l'usine du futur.} \emph{Dossier de
presse.}

Chahir, Mehdi, Bordel, Stéphanie, \& Somat, Alain (2022). Accompagner le
déploiement d'une nouvelle technologie par la prise en compte des
risques et des opportunités. \emph{Relations industrielles / Industrial
Relations}, \emph{77}(3). https://doi.org/10.7202/1094208ar

Cippelletti, Emma, Fournier, Etienne, \& Landry, Aurélie (2023, July
17-21). \emph{Acceptabilité des robots collaboratifs (Cobot) par des
travailleurs français.} In: A. Landry (ed.), Rendre les technologies
émergentes favorables à l'activité : expériences d'accompagnement de la
cobotique industrielle {[}Symposium{]}. 22\textsuperscript{e} congrès de
l'Association Internationale de Psychologie du Travail de Langue
Française (AIPTLF), Montréal, Québec.

Claverie, Bernard, Le Blanc, Benoît \& Fouillat, Pascal (2013). La
cobotique. \emph{Communication et Organisation}, \emph{44}, 203--214.
https://doi.org/10.4000/communicationorganisation.4425.

Clot, Yves, Faïta, Daniel, Fernandez, Gabriel, \& Scheller, Lisa (2000).
Entretiens en autoconfrontation croisée : une méthode en clinique de
l'activité. \emph{Perspectives interdisciplinaires sur le travail et la
santé}, (2-1). \uline{\url{https://doi.org/10.4000/pistes.3833}.}

Colgate, J. Edward, Wannasuphoprasit, Witaya, \& Peshkin, Michael A.
(1996). Cobots robots for collaboration with human operator. Proceedings
of the ASME Dynamic Systems and Control Division, 58, 433-440.

Colim, Ana, Faria, Carlos, Braga, Ana Cristina, Sousa, Nuno, Carneiro,
Paula, Costa, Nélson, \& Arezes, Pedro (2020). Towards an Ergonomic
Assessment Framework for Industrial Assembly Workstations - A Case Study
in \emph{Applied Science}, 10, 3048

Colim, Ana, Cardoso, André, Arezes, Pedro, Braga, Ana Cristina, Peixoto,
Ana Carolina, Peixoto, Vitor, Wolbert, Felix, Carneiro, Paula, Costa,
Nélson, \& Sousa, Nuno (2021). Digitalization of Musculoskeletal Risk
Assessment in a Robotic-Assisted Assembly Workstation. \emph{Safety},
7(4), Article 4. https://doi.org/10.3390/safety7040074

Colim, Ana, Faria, Carlos, Cunha, Joao, Oliveira, Joao, Sousa, Nuno, \&
Rocha, Luis A. (2021). Physical Ergonomic Improvement and Safe Design of
an Assembly Workstation through Collaborative Robotics. \emph{Safety},
7(1), Article 1. https://doi.org/10.3390/safety7010014

Coutarel Fabien, \& Daniellou François (2011) L'intervention ergonomique
pour la prévention des troubles musculosquelettiques\,: quels statuts
pour l'expérience et la subjectivité des travailleurs\,? \emph{Travail
Apprentissage}, 7(1), 62‑80.

David, G.C. (2005). Ergonomic Methods for Assessing Exposure to Risk
Factors for Work-Related Musculoskeletal Disorders. \emph{Occupational
Medicine}, 55, 190--199.

Devy, Michel (2012). La cobotique : des robots industriels aux robots
assistants, coopérants et co-opérateurs. \emph{Annales des
Mines-Réalités industrielles} (No. 1, pp. 76-85). Eska.
https://doi.org/10.3917/rindu.121.0076.

Dimitropoulos, Nikos, Togias, Theodoros, Zacharaki, Natalia, Michalos,
George, \& Makris, Sotiris (2021). Seamless human--robot collaborative
assembly using artificial intelligence and wearable devices.
\emph{Applied Sciences}, 11, 56-99.

Dubreuil-Nayrac, Isabelle, Caillat, Claire, Massu, Justine, Boisson,
Jean-François, \& Werlen, Martin (2019). Accompagner les transformations
des organisations. \emph{Le Journal des psychologues}, (5), 33-37.
https://doi.org/10.3917/jdp.367.0033.

EU-OSHA, (2019). Work-related musculoskeletal disorders: prevalence,
costs and demographics in the EU.
\url{https://osha.europa.eu/en/publications/summary-msds-facts-and-figures-overview-prevalence-costs-and-demographics-msds-europe/view}.

EUROGIP, (2017). Prévention dans le domaine de la robotique
collaborative. Synthèse de travaux réalisés à
l\textquotesingle international.
https://www.eurogip.fr/images/pdf/Prevention\%20robotique\%20collaborative\%20a\%20linternational-Eurogip129F.pdf.

Everaere, Christophe (2019). Autonomie. In: Gérard Valléry (ed.),
\emph{Psychologie du Travail et des Organisations : 110 notions clés}
(pp. 67-70). Paris: Dunod.
\url{https://doi10.3917/dunod.valle.2019.01.0067}

Ferreira Duarte, Nuno, Raković, Mirko, Tasevski, Jovica, Coco, Moreno,
Billard, Aude, \& Santos-Victor, José (2018). Action anticipation:
Reading the intentions of humans and robots. IEEE Robotics and
Automation Letters, 3(4), 4132-4139.

Fjell, Ylva, Alexanderson, Kristina, Karlqvist, Lena, \& Bildt, Carina,
(2007). Self-reported musculoskeletal pain and working conditions among
employees in the Swedish public sector. \emph{Work}, (28)33-46.

Forde, Martin S., Punnett, Laura, \& Wegman, David H. (2002).
Pathomechanisms of workrelated musculoskeletal disorders: Conceptual
issues. \emph{Ergonomics,} 45(9), 619--630.
https://doi.org/10.1080/00140130210153487.

Fournier, Etienne, Nstame Sima, Murielle, Jeoffrion, Christine, \&
Landry, Aurélie (2023, July 17-21). \emph{Comprendre les variabilités en
situation de travail avant l\textquotesingle implémentation
d\textquotesingle un robot collaboratif (Cobot)} In: A. Landry : Rendre
les technologies émergentes favorables à l'activité : expériences
d'accompagnement de la cobotique industrielle {[}Symposium{]}.
22\textsuperscript{e} congrès de l'Association Internationale de
Psychologie du Travail de Langue Française (AIPTLF), Montréal, Québec.

Galey, Louis, Terquem, Valérie, \& Barcellini, Flore (2022). A Social
Design Approach: Enhancement of Local Social Dialogue on the
Transformation of Work by Digital Technology. \emph{Relations
industrielles / Industrial Relations}, \emph{77}(3).
https://doi.org/10.7202/1094211ar

Gallagher, Sean, \& Schall Mark C. Jr. (2017). Musculoskeletal disorders
as a fatigue failure process: evidence, implications, and research
needs. \emph{Ergonomics}, 60(2), 255-269.
\href{https://doi.org/10.1080/00140139.2016.1208848}{\uline{https://doi.org/10.1080/00140139.2016.1208848}}.

\hspace{0pt}\hspace{0pt}Grenier-Pezé, Marie (2003). Corps et travail,
\emph{Cahiers du Genre}, 35 (2),141-164.

Groizeleau, Alain, Passerault, Jean-Michel, \& Esnard, Catherine (2019).
Méthode d'évaluation des risques psychosociaux auprès de travailleurs
déficients intellectuels. \emph{Psychologie du Travail et des
Organisations}, \emph{25}(3), 141-151.
\url{https://doi.org/10.1016/j.pto.2019.03.001}.

Guérin François, Pueyo Valérie, Béguin Pascal, Garrigou Alain, Hubault
François, Maline Jean, \& Morlet Thierry (2021). \emph{Concevoir le
travail, le défi de l'ergonomie}. Toulouse : Éditions Octares.

Guyon Taillens, Camille, Labrell, Florence, \& Demulier, Virginie
(2020). L'analyse des risques psychosociaux en ESAT parmi des
travailleurs en situation de handicap (TSH): éléments de réflexion.
Psychologie du Travail et des Organisations, 26(3), 239‑246.
\href{https://doi.org/10.1016/j.pto.2020.07.001}{\uline{https://doi.org/10.1016/j.pto.2020.07.001}}

France\textquotesingle s Institute for Research and Security. (2014).
\emph{Faits et chiffres en 2014}.
\url{http://www.inrs.fr/dms/inrs/CataloguePapier/ED/TI-ED-4458/ed4458.pdf}.

France\textquotesingle s Institute for Research and Security. (2015).
\emph{Troubles musculo-squelettiques : Facteurs de risque}.
\href{http://www.inrs.fr/risques/tms-troubles-musculosquelettiques/facteurs-risque.html}{\uline{http://www.inrs.fr/risques/tms-troubles-musculosquelettiques/facteurs-risque.html}}.

Haslam, Roger (2002). Targeting ergonomics interventions---learning from
health promotion. \emph{Applied Ergonomics}, 33, 241-249.
\uline{\url{https://doi.org/10.1016/S0003-6870(02)00016-9}.}

Hélardot, Valentine (2008). Pour une approche élargie de la pénibilité
du travail. \emph{Journal des professionnels de la santé au travail},
\emph{4}, 8-9.

Isaksen, Jesper (2000). Constructing meaning despite the drudgery of
repetitive work. \emph{Journal of humanistic Psychology}, 40(3), 84-107.
https://doi.org/10.1177/0022167800403008.

\hypertarget{lop-nor-suzila-2011.-a-literature-review-of-ergonomics-risk-factors-in-construction-industry.-procedia-engineering-20-89-97.}{%
\texorpdfstring{\& Lop, Nor Suzila (2011). A literature review
of ergonomics risk factors in construction industry. \emph{Procedia
Engineering, 20, 89-97.
}{\& Lop, Nor Suzila (2011). A literature review of ergonomics risk factors in construction industry. Procedia Engineering, 20, 89-97. }}\label{lop-nor-suzila-2011.-a-literature-review-of-ergonomics-risk-factors-in-construction-industry.-procedia-engineering-20-89-97.}}

Jafflin, Aurélie, \& Nadeau, Sylvie (2020). Identifying OHS hazards in
university research laboratories from an ergonomics and human factors
perspective. Paper presented at the 66th Gfa-frühjahrs Kongress, Berlin,
Germany.

Jansen, Anne, Beek, Dolf Van Der, Cremers, Anita, Neerincx, Mark, \&
Middelaar, Johan Van (2018). Emergent risks to workplace safety; working
in the same space as a cobot.
https://publications.tno.nl/publication/34627026/je8DYe/TNO-2018-R10742.pdf.

Kagermann, Henning, Helbig, Johannes, Hellinger, Ariane, \& Wahlster,
Wolfgang (2013). \emph{Recommendations for implementing the strategic
initiative INDUSTRIE 4.0: Securing the future of German manufacturing
industry; final report of the Industrie 4.0 Working Group}.
Forschungsunion.
https://en.acatech.de/publication/recommendations-for-implementing-the-strategic-initiative-industrie-4-0-final-report-of-the-industrie-4-0-working-group/.

Karsh, Ben Tzion, Moro, Franscico B.P., \& Smith, Michael J. (2001). The
efficacy of workplace ergonomic interventions to control musculoskeletal
disorders: a critical analysis of the peer-reviewed literature.
\emph{Theoretical Issues in Ergonomics Science,} 2 (1), 23--96.
https://doi.org/10.1080/14639220152644533.

Kim, Wansoo, Peternel, Luka, Lorenzini, Marta, Babič, Jan, \& Ajoudani,
Arash (2021) A Human-Robot Collaboration Framework for Improving
Ergonomics During Dexterous Operation of Power Tools. \emph{Robotics and
Computer-Integrated Manufacturing}, 68.

Kuorinka, Likka, \& Forcier, Lina (1995). \emph{Work related
musculoskeletal disorders (WMSDs): A reference book for prevention.}
London: Taylor and Francis.

Landry, Aurélie (2012) Suggested Evaluation Approach for Training in
Ergonomics, \emph{Work: A Journal of Prevention Assessment, and
Rehabilitation.} 41 (2), 177-186. \uline{doi: 10.3233/WOR-2012-1282}

Lanfranchi, Jean-Baptiste, \& Duveau, Aurélie (2008). Explicative models
of musculoskeletal disorders (MSD): From biomechanical and psychosocial
factors to clinical analysis of ergonomics. \emph{European Review of
Applied Psychology}, 58(4), 201‑213.
\href{https://doi.org/10.1016/j.erap.2008.09.004}{\uline{https://doi.org/10.1016/j.erap.2008.09.004}}

Lasfargues, Gérard, Molinie, Anne-Françoise, \& Volkoff, Serge (2005).
Départs en retraite et « travaux pénibles ». \emph{L'usage des
connaissances scientifiques sur le travail et ses risques à long terme
pour la santé, Rapport de recherches}, (19), 39.

LeBreton, J. M., \& Senter, J. L. (2008). Answers to 20 Questions about
Interrater Reliability and Interrater Agreement. \emph{Organizational
Research Methods}, 11, 815-852.
http://dx.doi.org/10.1177/1094428106296642

Leclerc, Annette, Chastang, Jean-François, Pascal, P., Plouvier,
Sandrine, \& Mediouni, Zakia (2015). Conséquences des troubles
musculosquelettiques sur l'itinéraire professionnel, résultats d'une
enquête nationale. \emph{Archives des Maladies Professionnelles et de
l\textquotesingle Environnement}, 76(3, 245-54.
https://doi.org/10.1016/j.admp.2014.10.009.

Légifrance (2020). Article L5213-13 du code du travail : Sous-section 3
: Entreprises adaptées et centres de distribution de travail à domicile.
\href{https://www.legifrance.gouv.fr/affichCode.do?idSectionTA=LEGISCTA000006195890\&cidTexte=LEGITEXT000006072050}{\uline{https://www.legifrance.gouv.fr/affichCode.do?idSectionTA=LEGISCTA000006195890\&cidTexte=LEGITEXT000006072050}}.

Leplat, Jean (1989). Error analysis, instrument and object of task
analysis. \emph{Ergonomics, 32}(7), 813-822.

Levanon, Yafa, Gefen, Amit, Lerman, Yehuda, Givon, Uri, \& Ratzon, Navah
Z. (2012). Reducing musculoskeletal disorders among computer operators:
Comparison between ergonomics interventions at the workplace.
\emph{Ergonomics}, 55 (12), 1571-85.
\href{https://doi.org/10.1080/00140139.2012.726654}{\uline{https://doi.org/10.1080/00140139.2012.726654}}.

Lorenzini, Marta, Kim, Wansu, Momi, Elena de, \& Ajoudani, Arash (2019).
A New Overloading Fatigue Model for Ergonomic Risk Assessment with
Application to Human-Robot Collaboration. In Proceedings of the 2019
\emph{International Conference on Robotics and Automation} (ICRA),
Montreal, QC, Canada, 20--24 May 2019; pp. 1962--1968.

Lu, Yang (2017). Industry 4.0: A survey on technologies, applications
and open research issues. \emph{Journal of Industrial Information
Integration}, 6, 1-10. \url{https://doi.org/10.1016/j.jii.2017.04.005}.

Morin, Estelle (2008). Sens du travail, santé mentale au travail et
engagement organisationnel. \emph{Cahier de recherche}, \emph{543},
99-193.

Moulières-Seban, Théo, Bitonneau, David, Salotti Jean-Marc, Thibault,
Jean-François, \& Claverie, Bernard (2017). Human Factors Issues for the
Design of a Cobotic System. In: P., Savage-Knepshield \& J. Chen (Eds),
\emph{Advances in Human Factors in Robots and Unmanned Systems}.
Advances in Intelligent Systems and Computing (375-385). Springer, Cham.

MTEFPDS (Department of Employment, Social Affairs \& Inclusion) (2015).
\emph{Conditions de travail. Bilan 2015}.
\url{https://travail-emploi.gouv.fr/IMG/pdf/bilan_des_conditions_de_travail_2015.pdf}.

National Research Council. (2001). \emph{Musculoskeletal Disorders and
the Workplace}. Washington, DC: National Academy Press.

Nunes, Isabel (2009). FAST ERGO\_X--a tool for ergonomic auditing and
work-related musculoskeletal disorders prevention. \emph{Work},
\emph{34}(2), 133-148.

Nunes, Isabel., \& McCauley-Bush, Pamela (2012). Work-related
musculoskeletal disorders assessment and prevention. In: Nunes, I. L.
(Ed.), \emph{Ergonomics-A Systems Approach}, InTech (pp.1--30).
http://www.intechopen.com/books/ergonomics-a-systemsapproach/work-related-musculoskeletal-disorders-assessment-and-prevention.

Roquelaure, Yves (2018). Musculoskeletal disorders and psychosocial
factors at work. ETUI Research Paper-Report, 142.
https://www.etui.org/content/download/35699/355072/file/EN-Report-142-MSD-Roquelaure-WEB.pdf.

Roquelaure, Yves, Bodin, Julie, Ha, Catherine, Petit Le Manac'h, Audrey,
Descatha, Alexis, Leclerc, Annette, Goldberg, Marcel, Imbernon, Ellen
(2011). Personal, biomechanical, and psychosocial risk factors for
rotator cuff syndrome in a working population. \emph{Scandinavian
Journal of Work and Environmental Health}, 37(6), 502-511. Doi
:10.5271/sjweh.3179.

Sérazin, Céline, Ha, Catherine, Bodin, Julie, Imbernon, Ellen, \&
Roquelaure, Yves. (2013). Employment and occupational outcomes of
workers with musculoskeletal pain in a French region. \emph{Occupational
and Environmental Medicine,} 70, 143-148.
\href{http://dx.doi.org/10.1136/oemed-2012-100685}{\uline{http://dx.doi.org/10.1136/oemed-2012-100685}}.

Shafti, Ali, Ataka, Ahmad, Lazpita, Beatriz, Shiva, Ali, Wurdemann,
Helge, \& Althoefer, Kaspar (2019). Real-Time Robot-Assisted Ergonomics.
In: Proceedings of the 2019 \emph{International Conference on Robotics
and Automation} (ICRA), Montreal, QC, Canada, 20--24 May 2019; pp.
1975--1981.

St Vincent, Marie, Lortie, Monique, \& Chicoine, Denise (2001).
Participatory Ergonomics Training in the Manufacturing Sector and
Ergonomic Analysis Tools. \emph{Relations industrielles / Industrial
Relations}, \emph{56}(3), 491--515.
\href{https://doi.org/10.7202/000080ar}{\uline{https://doi.org/10.7202/000080ar}}

St Vincent, Marie, Vézina, Nicole, Bellemare, Marie, Denys, Denis,
Ledoux, Elise, \& Imbeau, Daniel (2011). \emph{L'intervention en
ergonomie}. Québec : Éditions Multi-Mondes et Institut de recherche
Robert-Sauvé en santé et sécurité du travail (IRSST), 2011, 360 p., ISBN
: 978-2-89544-165-6.

Thames, Lane, \& Schaefer, Dirck (2016). Software-defined cloud
manufacturing for Industry 4.0. \emph{Procedia CIRP}, 52, 12-17.
\href{https://doi.org/10.1016/j.procir.2016.07.041}{\uline{https://doi.org/10.1016/j.procir.2016.07.041}}

Theureau, Jacques (2004). L'hypothèse de la cognition (ou action) située
et la tradition d'analyse du travail de l'ergonomie de langue française.
\emph{Activités}, 01(2), Article 2.
https://doi.org/10.4000/activites.1219

Tinsley, Howard, \& Weiss, David (1975). Interrater reliability and
agreement of subjective judgments. \emph{Journal of Counseling
Psychology, 22}(4), 358--376.
\href{https://psycnet.apa.org/doi/10.1037/h0076640}{https://doi.org/10.1037/h0076640}

Trevelyan, F., \& Haslam, Roger (2001). Musculoskeletal disorders in a
handmade brick manufacturing plant. \emph{International Journal of
Industrial Ergonomics,} 27, 43-55.
\url{https://doi.org/10.1016/S0169-8141(00)00036-6}.

Vallery, Gérard, Bobillier Chaumon, Marc-Eric, Brangier, Eric, \&
Dubois, Michel (2019). \emph{Psychologie du Travail et des Organisations
: 110 notions clés-2e éd}. Dunod.

Villani, Valeria, Pini, Fabio, Leali, Franscesco, \& Secchi, Cristian
(2018). Survey on human--robot collaboration in industrial settings\,:
Safety, intuitive interfaces and applications. \emph{Mechatronics}, 55,
248‑266. https://doi.org/10.1016/j.mechatronics.2018.02.009

Whysall, Zara, Haslam, Roger, \& Haslam, Cheryl (2004). Processes,
barriers, and outcomes described by ergonomics consultants in preventing
work-related musculoskeletal disorders. \emph{Applied Ergonomic}s,
35(4), 343-351. https://doi.org/10.1016/j.apergo.2004.03.001.

Whysall, Zara, Haslam, Cheryl, \& Haslam, Roger (2006). Implementing
health and safety interventions in the workplace: An exploratory study.
\emph{International Journal of Industrial Ergonomics}, \emph{36}(9),
809-818. https://doi.org/10.1016/j.ergon.2006.06.007.

World Health Organization (2003). Preventing musculoskeletal disorders
in the workplace. Retrieved from
https://apps.who.int/iris/bitstream/handle/10665/42651/924159053X.pdf?sequence=1.

World Health Organization (2019). Musculoskeletal conditions. Retrieved
from
https://www.who.int/news-room/fact-sheets/detail/musculoskeletal-conditions.

Xu, Yan-Wen, Cheng, Andy, \& Li-Tsang, Cecilia (2013). Prevalence and
risk factors of work-related musculoskeletal disorders in the catering
industry: A systematic review. \emph{Work}, \emph{44}(2), 107-116.

\end{document}